\documentclass[conference]{IEEEtran}

\usepackage[utf8]{inputenc}
\usepackage{amsmath} 
\usepackage{amssymb}  
\usepackage{amsfonts}
\usepackage{mathrsfs}
\usepackage{xspace}
\usepackage{xfrac}
\usepackage[compress]{cite}
\usepackage{balance}
\usepackage{color}
\usepackage[footnotesize]{caption}
\usepackage{subfig}

\usepackage{pgfplots}
\usepackage{tikz}
\tikzset{every picture/.append style={font=\scriptsize}}
\usetikzlibrary{external,calc}
\tikzset{%
  >=latex, 
  inner sep=0pt,%
  outer sep=2pt,%
  mark coordinate/.style={inner sep=0pt,outer sep=0pt,minimum size=3pt,
    fill=black,circle}%
}

\usepackage{graphicx}
\graphicspath{{figures/}}

\newcommand{\R}{{\mathbb R}}

\renewcommand{\H}{{\mathcal{H}}}

\newcommand{\0}{\boldsymbol{0}}
\newcommand{\I}{\mathbf{I}\xspace}

\renewcommand{\tt}{\boldsymbol{t}}

\newcommand{\G}{\text{G}\xspace}
\newcommand{\F}{\mathcal{F}\xspace}
\newcommand{\g}{\mathfrak{g}\xspace}

\newcommand{\SO}{\text{SO}\xspace}
\newcommand{\so}{\mathfrak{so}\xspace}

\newcommand{\x}{\boldsymbol{x}\xspace}

\DeclareMathOperator{\ad}{ad}
\DeclareMathOperator{\Ad}{Ad}

\DeclareMathOperator{\mmod}{mod}

\DeclareMathOperator{\wPi}{w^{\pi}}

\begin{document}

%
\title{On wrapping the Kalman filter and estimating with the SO(2) group}

\author{\IEEEauthorblockN{Ivan Marković, Josip Ćesić, Ivan Petrović}
\IEEEauthorblockA{University of Zagreb, Faculty of Electrical Engineering and Computing, Unska 3, 10000 Zagreb, Croatia\\
Email: \texttt{ivan.markovic@fer.hr, josip.cesic@fer.hr, ivan.petrovic@fer.hr}}
}

\maketitle

\begin{abstract}
This paper analyzes directional tracking in 2D with the extended Kalman filter on Lie
groups (LG-EKF).
The study stems from the problem of tracking objects moving in 2D Euclidean space, with the observer measuring
direction only, thus rendering the measurement space and object position on the circle---a non-Euclidean geometry.
The problem is further inconvenienced if we need to include higher-order dynamics in the state space,
like angular velocity which is a Euclidean variables. 
The LG-EKF offers a solution to this issue by modeling the state space as a Lie group or combination thereof, e.g.,
$\SO(2)$ or its combinations with $\R^n$.
In the present paper, we first derive the LG-EKF on $\SO(2)$ and subsequently show that this derivation, based on the
mathematically grounded framework of filtering on Lie groups, yields the same result as heuristically wrapping the
angular variable within the EKF framework.
This result \emph{applies only} to the $\SO(2)$ and $\SO(2)\times\R^n$ LG-EKFs and is not intended to be extended to
other Lie groups or combinations thereof.
In the end, we showcase the $\SO(2)\times\R^2$ LG-EKF, as an example of a constant angular acceleration model, on the
problem of speaker tracking with a microphone array for which real-world experiments are conducted
and accuracy is evaluated with ground truth data obtained by a motion capture system.
\end{abstract}


\IEEEpeerreviewmaketitle

\section{Introduction}\label{sec:intro}

In moving object tracking, it is not uncommon to work with sensors that can provide only direction to the object in
question.
The measurement and estimation state space have a specific geometry of their own, which is different from the geometry of
the true trajectory space.
The problem is challenging, because, although the motion of the object resides either in 3D or 2D Euclidean space,
corresponding measurements reside either on the sphere or the circle, respectively. 
Namely, if we are measuring and estimating only the direction to the object in 2D, i.e., the azimuth, the state
and measurements will bear the non-Euclidean properties of angles.
However, if we are to extend the state space so that it includes both the angular velocity and acceleration (which are
Euclidean variables), so that we can apply a higher-order dynamic motion model, we are faced with constructing a
`hybrid' state space consisting of both the non-Euclidean and Euclidean variables.

There exist Bayesian methods based on the principle of assumed density filtering with directional distributions on the
circle, namely the von Mises distribution, the wrapped Gaussian distribution and the Bingham distribution (which
actually models variables with $180^{\circ}$ symmetry), that capture intrinsically the non-Euclidean nature of angular
random variables \cite{Azmani2009a,Markovic2010a,Markovic2012a,Kurz2013a,Kurz2013c,Stienne2014}.
The benefit of these approaches is that they take globally into account the geometry of the state space.
For example, in the case of the von Mises distribution it has been shown that the filter outperforms the naive Kalman filter, which treats angles like regular Euclidean variables, and the modified Kalman filter, which takes into account the nature of angles by wrapping them on the circle \cite{Kurz2013a,Markovic2015}.
However, extending the state space with additional variables of different geometry, e.g., to analytically model the azimuth with the von Mises distribution and the range or the angular velocity with the Gaussian distribution and capture correctly the cross-correlations, remains a challenge.


The $\SO(2)$ group is a set of orthogonal matrices with determinant one, whose elements geometrically
represent rotations.
This makes it an interesting candidate for estimation with angular variables.
Furthermore, a filter could be derived not just for $\SO(2)$, but also for combinations of $\SO(2)$ with $\R$.
This would enable us to create the aforementioned `hybrid' state vector that would join both non-Euclidean and Euclidean
variables within the same filter and enable a seamless utilization of higher-order system models with constant angular
velocity or acceleration.
An extended Kalman filter on matrix Lie groups was recently proposed in \cite{Bourmaud2014}.
It provides us with a mathematical framework for solving the `hybrid' state space problem.
Indeed, the filter can be applied directly for any state that is a combination of Lie groups, since a Cartesian product
of Lie groups is a Lie group \cite{Bourmaud2014}.
However, it should be noted that the LG-EKF is a local approach, in the sense that it does not take globally the geometry of the state space into account, but locally captures the geometry of state space via exponential mapping.
Another approach would be to model the whole state space as a Euclidean vector within the `classical' Kalman filter framework, and wrap the operations involving angular variables.
Indeed, this was performed in \cite{Kurz2013a} to modify the unscented Kalman filter for angular state estimation, in \cite{Crouse2015} to take idiosyncrasies of directional statistics when using polar or spherical coordinates in the cubature Kalman filter, and in \cite{Markovic2015} to modify the Gaussian mixture probabilistic hypothesis density filter for multitarget tracking on a circle.


In this paper we propose to analyze the LG-EKF for directional tracking of moving objects in 2D.
First, we look into deriving the LG-EKF on the $\SO(2)$, which also serves as a gentle introduction to the subject matter since the LG-EKF introduces non-trivial notation.
Second, we model the directional moving object tracking in 2D as an estimation problem on the
Lie group composed of the direct product $\SO(2)\times\R^2$, i.e., a group that represents the moving object
azimuth, angular velocity and angular acceleration.
For the motion model, we use the constant angular acceleration model.
In the end, we show that the $\SO(2)$ LG-EKF filter derivation based on the mathematically grounded framework of
filtering on Lie groups yields the same result as heuristically wrapping the angular variables within the
extended Kalman filter (EKF) framework.
Since for the case of $\R^n$ the LG-EKF evaluates to EKF \cite{Bourmaud2014}, this results also extends to
$\SO(2)\times\R^2$ LG-EKF and an $\R^3$ EKF when wrapping the angular component.
Please note that this result \emph{applies only} to the $\SO(2)$ filter and is not intended to be extended to other
Lie groups or combinations thereof.
Indeed, given that $\SO(2)$ is abelian, i.e., commutative, the result does
not seem unexpected, but we assert that it gives interesting theoretical perspective on estimation and tracking with the
heuristically modified EKF.
Before we proceed with the filter derivation, we introduce some necessary formal definitions and operators
for working with matrix Lie groups.

\section{Mathematical background}

\subsection{Wrapping the Kalman filter}

In this section we shall assume that wrapping operation amounts to enforcing the angular variable to be in the $[-\pi, \pi]$ interval, and we designate this operation as follows
\begin{equation}\label{eq:wrapToPi}
  \wPi(x) = \mmod(x + \pi, 2\pi) - \pi.
\end{equation}
%
Note that when computing the difference between two angular variables, the wrapping effect of the circle should be taken
into account, e.g., the difference between $178^{\circ}$ and $-178^{\circ}$ should evaluate to $4^{\circ}$.
This is also achieved by \eqref{eq:wrapToPi} when the difference is given as the argument, i.e., difference between two
angles $x$ and $y$ is computed as $\wPi(x-y)$.

Let us assume the following system model
\begin{equation}\label{eq:kf_state_prediction}
  x_{k+1|k} = f_k(x_k, u_k) + n_k, \quad n_k \sim \mathcal{N} (0, Q)
\end{equation}
where $x_k$ is the system state, $u_k$ is the control input, $n_k$ is process noise, and $f_k(\,\cdot\,)$ is the
non-linear system state equation.
In the EKF the idiosyncrasies of angular data appear most prominently in the correction step when
calculating the innovation, which should be computed as
\begin{equation}\label{eq:kf_state_correction}
  x_{k+1} = x_{k+1|k} + K_k \wPi(z_k - h_k(x_{k+1|k})), 
\end{equation}
where $K_k$ is the Kalman gain, $z_k$ is the measurement, and $h_k(\,\cdot\,)$ is the non-linear measurement equation.

To demonstrate this, let us take a simple example of having an identity measurement equation, $x_{k+1|k} = 358^{\circ}$,
$z_k = 2^{\circ}$ and $K_k = 0.5$.
If we would not wrap the innovation, the updated state would yield a clearly incorrect result of $x_{k+1} =
180^{\circ}$ inlieu of $x_{k+1} = 360^{\circ}$.
For practical purposes, after correction and prediction the system state can be checked to the required interval by
computing $x_{k+1} \leftarrow \wPi(x_{k+1})$.
In the sequel when we refer to the modified Kalman filter, it entails treating angular variables with the previously
introduced operation.
Furthermore, we assume that the reader is familiar with EKF equations, which we will not present or derive explicitly
in order to keep the brevity of the paper.

\subsection{Lie Groups}\label{sec:filtering_matrix_lie_groups}


A Lie group $\G$ is a group which is also a smooth manifold and the group composition and inverse are smooth
functions on the manifold $\G$.
A manifold is an object that looks locally like a piece of $\R^n$ and $\G$ is `smooth' in the sense that is has a tangent
space, of the appropriate dimension, at each point.
Take for example the circle, a curve in $\R^2$ which looks locally (but not globally) like $\R^1$.
For a matrix Lie group the composition and inverse are simply matrix multiplication and inversion, with the
identity element $\I^{n \times n}$ \cite{Chirikjian2012b}.

A Lie Algebra $\g$ is an open neighborhood of $\0^{n \times n}$ in the tangent space of
$\G$ at the identity $\I^{n \times n}$.
The matrix exponential $\exp_{\G}$ and matrix logarithm $\log_{\G}$ establish a local diffeomorphism between Lie groups
and Lie algebras \cite{Bourmaud2014}
\begin{align}
  \exp_{\G} : \g \rightarrow \G, \quad
  \log_{\G} : \G \rightarrow \g.
\end{align}
The Lie Algebra $\g$ associated to a $p$-dimensional matrix Lie group $\G \subset \R^{n \times n}$ is a $p$-dimensional
vector space%
\cite{Chirikjian2012b}.
A linear isomorphism between $\g$ and $\R^p$ is given by
\begin{align}
  [\cdot]^{\vee}_{\G} : \g \rightarrow \R^p, \quad
  [\cdot]^{\wedge}_{\G} : \R^p \rightarrow \g.
\end{align}
Lie Groups are not necessarily commutative.
The following two operators capture this property
\begin{itemize}
  \item the adjoint representation of $\G$ on $\R^p$
    \begin{align}
      \Ad_{\G} : \Ad_{\G}(X) x =
      \left[ X [x]^{\wedge}_{\G} X^{-1} \right]^{\vee}_{\G}
    \end{align}
  \item the adjoint representation of $\R^p$ on $\R^p$
    \begin{align}
      \ad_{\G} : \ad_{\G}(x) y =
      \left[ [x]^{\wedge}_{\G} [y]^{\wedge}_{\G} - [y]^{\wedge}_{\G} [x]^{\wedge}_{\G}\right]^{\vee}_{\G}
    \end{align}
\end{itemize}
where $x,y\in \R^p$.
In the sequel, these operators, the exponential and logarithmic mapping are given concrete form for the pertinent
 Lie groups.

\subsection{The $\SO(2)$ group}



In this example our system state (azimuth of the tracked object) is modeled as 
the group $G=\SO(2)$, i.e., as the rotation matrix $X_k=R_{\theta_k}$
\begin{equation}
  R_{\theta_k} = 
  \begin{bmatrix}
    \cos\theta_k & -\sin\theta_k\\
    \sin\theta_k & \cos\theta_k
  \end{bmatrix}.
\end{equation}
The composition and inverse in $\SO(2)$ are simply evaluated as $X_1X_2 = R_1R_2, X^{-1} = R^{\text{T}}$.
For this case the associated Lie algebra which bridges $X_k\in\G$ and $x_k = \theta_k\in\R^1$ is $\g=\so(2)$, and the following
holds
\begin{equation}\label{eq:so2_hat_operator}
  [\theta_k]^{\wedge}_{\G} = 
  \begin{bmatrix}
     0 & -\theta_k\\
     \theta_k & 0
  \end{bmatrix}.
\end{equation}
%
%
The link between $\SO(2)$ and $\so(2)$ is given by the exponential and logarithmic mapping
\begin{align}
  &\exp_{\G}([\theta_k]^{\wedge}_{\G})
  =
  R_{\theta_k} \,:\, \so(2)\rightarrow\SO(2),\\
  &\log_{\G}(R_{\theta_k})
  =
  [\theta_k]^{\wedge}_{\G}\, : \,\SO(2)\rightarrow\so(2).
\end{align}
Due to the commutativity of $\SO(2)$, the adjoint operators are
\begin{equation}\label{eq:adjoint_operators}
  \ad_{\G}(\theta_k) = 0, \quad \Ad_{\G}\left( \exp_{\G} \left( [\theta_k]^{\wedge}_{\G} \right) \right) =
  1.
\end{equation}
These properties greatly simplify the LG-EKG formulae for the $\SO(2)$ group which will become evident in the sequel.

\subsection{The $\SO(2)\times \R^2$ group}

\noindent In this section we propose to model the system state as the Cartesian
product of groups $\G=\SO(2)\times\R^2$.
This is a slight abuse of notation intended for clarity, since when talking about $\R$ within the group or algebra,
we are actually referring to the group of algebra representation of $\R$, for which the explicit representation is given
further in the paper.
The moving object state $X_k$ will represent the azimuth of the target as a rotation matrix
$R_{\theta_k}\in\SO(2)$, angular velocity as a real number $\omega_k\in\R$, and angular acceleration also as a real number
$\alpha_k\in\R$.
The system state $X_k$ can be symbolically represented as
\begin{align}
  X_k = 
  \begin{bmatrix}
    R_k & & \\
    & 
    \begin{bmatrix}
      1 & \omega_k\\
      0 & 1
    \end{bmatrix}
    & \\
    & & 
    \begin{bmatrix}
      1 & \alpha_k\\
      0 & 1
    \end{bmatrix}
  \end{bmatrix}
  =
  \begin{pmatrix}
    R_k\\
    \omega_k\\
    \alpha_k
  \end{pmatrix}_{\G}  \,.
\end{align}
Note that composition and inverse on such a group is evaluated as follows
\begin{equation}
  X_1X_2 =
  \begin{pmatrix}
    R_1R_2\\
    \omega_1 + \omega_2\\
    \alpha_1 + \alpha_2
  \end{pmatrix}_{\G},
  \quad
  X^{-1} = 
  \begin{pmatrix}
    R^{\text{T}}\\
    -\omega\\
    -\alpha
  \end{pmatrix}_{\G}.
\end{equation}

The associated Lie algebra is $\g=\so(2)\times\R^2$ which bridges the state on the Lie group $X_k \in \G$ with the
vector $x_k = [\theta_k \ \omega_k \ \alpha_k]^{\text{T}} \in \R^3$, and the following holds
\begin{align}
  [x_k]^{\wedge}_{\G} = 
  \begin{bmatrix}
    [\theta_k]^{\wedge}_{\SO(2)} & & \\
    & 
    [\omega_k]^{\wedge}_{\R}
    & \\
    & & 
    [\alpha_k]^{\wedge}_{\R}
  \end{bmatrix}
  =
  \begin{pmatrix}
    [\theta_k]^{\wedge}_{\SO(2)}\\
    \omega_k\\
    \alpha_k
  \end{pmatrix}_{\g}  \,,
\end{align}
where $[\theta_k]^{\wedge}_{\SO(2)}$ is given by \eqref{eq:so2_hat_operator}, while
\begin{equation}
  \begin{split}
  [\omega_k]^{\wedge}_{\R} &= 
  \begin{bmatrix}
     0 & \omega_k\\
     0 & 0
  \end{bmatrix}
  \quad \text{and} \quad
  [\alpha_k]^{\wedge}_{\R} = 
  \begin{bmatrix}
     0 & \alpha_k\\
     0 & 0
  \end{bmatrix}.
  \end{split}
\end{equation}
The link between the group $\G$ and the associated algebra $\g$ is defined by the exponential mapping
\begin{align}
  \exp_{\G}\left( [x_k]^{\wedge}_{\G} \right)
  &=
  \begin{pmatrix}
    \exp_{\SO(2)}\left( [\theta_k]^{\wedge}_{\SO(2)} \right)\\
    \omega_k\\
    \alpha_k
  \end{pmatrix}_{\G}
  =
  \begin{pmatrix}
    R_k\\
    \omega_k\\
    \alpha_k
  \end{pmatrix}_{\G},
\end{align}
and logarithmic mapping
\begin{align}\label{eq:log_mapping}
  \log_{\G}\left( X_k \right)
  =
  \begin{pmatrix}
    \log_{\SO(2)}\left(R_k\right)\\
    \omega_k\\
    \alpha_k
  \end{pmatrix}_{\g}
  =
  \begin{pmatrix}
    [\theta_k]^{\wedge}_{\SO(2)}\\
    \omega_k\\
    \alpha_k
  \end{pmatrix}_{\g}.
\end{align}
%
Furthermore, since $\SO(2)$ and $\R$ are abelian and the Cartesian product of abelian groups is abelian, the adjoint
operators are again trivial
\begin{equation}\label{eq:adjoint_operators_so2rr}
  \ad_{\G}(x_k) = \0^{3\times 3}, \quad \Ad_{\G}\left( \exp_{\G} \left( [x_k]^{\wedge}_{\G} \right) \right) =
  \I^{3\times 3}.
\end{equation}

\section{The EKF on Matrix Lie Groups}

As in the case of classical Kalman filtering, we need to begin by defining a motion model by which we will
calculate the prediction.
For general filtering on matrix Lie groups, the system model is defined by the following equation \cite{Bourmaud2014}
\begin{equation}\label{eq:lgekf_system_model}
  \begin{split}
   X_{k+1}
    = f(X_{k},u_k, n_{k})
    = X_{k} \, \exp_{\G} \left( [\hat{\Omega}_k + n_{k}]^{\wedge}_{\G} \right) \,,
  \end{split}
\end{equation}
where $X_k \in \G$ is the system state at time $k$, $\G$ is a $p$-dimensional Lie Group, $n_k \sim
\mathcal{N}_{\R^p}(\0^{p \times 1},Q_k)$ is white Gaussian noise and $\hat{\Omega}_k=\Omega(X_{k},u_k):\G\times\R^w \rightarrow
\R^p$ is the system state equation which describes how the model acts on the state and control input in order to calculate the displacement.

Note that the function of $\hat{\Omega}_k$ is to take the system state which resides on \G and the control input which resides on $\R^w$, calculate the displacement
by applying the system model, and then transfer the displacement to the vector space $\R^p$ where additive noise is
added.
This displacement is then transferred to the associated Lie algebra by the $[\,.\,]^{\wedge}_{\G}$ operator and then
exponentially mapped back to the Lie group to be added by way of composition to the system state $X_k$.
Given that, a question arises how to implement a specific system model, since in LG-EKF it operates through a
displacement?
That is, how to construct $\hat{\Omega}_k$ from $f_k(x_k, u_k)$?
The first step would be to write the system equation as $f_k(x_k, u_k) = x_k + \hat{f}_k(x_k, u_k)$ which can then be
practically `translated' to appropriate $\hat{\Omega}_k$.
Note that generality is not lost here since $-x_k$ can be included within $\hat{f}_k(x_k, u_k)$, 

The prediction step of the LG-EKF is governed by the following formulae \cite{Bourmaud2014}
\begin{align}
  \mu_{k+1|k} & = \mu_{k} \exp_{\G}  \left( [\hat{\Omega}_k]^{\wedge}_{\G} \right)\label{eq:lgekf_mu_prediction}\\
  P_{k+1|k} & = \F_{k} P_k \F_{k}^T + \Phi_{\G}(\hat{\Omega}_{k}) Q_k \Phi_{\G}(\hat{\Omega}_{k})^T
  \,,\label{eq:lgekf_cov_prediction}
\end{align}
where $\mu_{k}\in\G$ is the estimated mean value of the system state $X_k$, $P_k\in\R^{p\times p}$ is the estimated
covariance matrix, while other terms are non-trivially calculated matrices
\begin{align}
  \F_k & = \text{Ad}_{\G} \left( \exp_{\G} \left([-\hat{\Omega}_k]^{\wedge}_{\G} \right) \right) +
  \Phi_{\G}(\hat{\Omega}_{k}) \mathscr{C}_{k}\label{eq:lgekf_Fk_matrix},\\
  \Phi_{\G}({\nu}) &= \sum_{m=0}^{\infty} \frac{(-1)^m}{(m+1)!}\ad_{\G}( {\nu}
  )^m, \quad {\nu}\in\R^{p}\label{eq:lgekf_PhiG_matrix},\\
  \mathscr{C}_{k} & = \dfrac{\partial}{\partial \epsilon} \Omega \left( \mu_{k} \exp_{\G} \left(
  [\epsilon]^{\wedge}_{\G} \right), u_{k-1} \right)_{|\epsilon=0}\,.\label{eq:lgekf_Ck_matrix}
\end{align}
The parameter $\epsilon\in\R^p$ can be seen as a \emph{Lie algebraic error} which is approximated as being distributed
according to a classical Euclidean Gaussian distribution $\epsilon \sim \mathcal{N}_{\R^p}(\0^{p\times 1}, P_k)$.
It is interesting to note that the mean value $\mu_k$ resides on the Lie group $\G$, while the covariance matrix $P_k$ 
describes uncertainty in $\R^p$.
Although at first this appears peculiar, it is a consequence of modeling the uncertainty of states on
Lie groups by the assumption of the concentrated Gaussian distribution $X_k \sim \mathcal{G}(\mu_k, P_k)$.
In essence, the state resides on the group, but its uncertainty resides on the tangential vector space.
For a more formal introduction of this concept, please confer \cite{Bourmaud2014}.


The discrete measurement model on the matrix Lie Group is given as follows
\begin{align}
  z_{k+1} = h(X_{k+1}) \, \exp_{\G'} \left( [m_{k+1}]^{\wedge}_{\G'} \right) \,,
\end{align}
where $z_{k+1} \in \G'$, $h:\G \rightarrow \G'$, and $m_{k+1}\sim
\mathcal{N}_{\R^q}(\0^{q \times 1},R_k)$ is white Gaussian noise.
Note that here a different group $\G'$ is used since the system state and measurements might belong to different groups.
Having the measurement model defined, we can proceed now to the update step which will first constitute the calculation
of the Kalman gain
\begin{align}
  K_{k+1} = P_{k+1|k} \H_{k+1}^T \left( \H_{k+1} P_{k+1|k} \H_{k+1}^T + R_{k+1} \right)^{-1} \,,
\end{align}
where the measurement matrix $\H_{k+1}$ is calculated via
\begin{equation}\label{eq:calH}
  \begin{split}
    \H_{k+1} = \dfrac{\partial}{\partial \epsilon}
    &\left[
    \log_{\G'} 
    \left(
    h(\mu_{k+1|k})^{-1}\right.\right.\\
    &\left.\left. h \left( \mu_{k+1|k} \exp_{\G}  \left( [\epsilon]^{\wedge}_{\G} \right)
    \right)
    \right)
    \right]^{\vee}_{\G \ | \epsilon=0} \,.
    \end{split}
\end{equation}
Furthermore, the innovation vector multiplied by Kalman gain is computed as
\begin{align}
  \nu_{k+1} = 
  K_{k+1}
  \left[
  \log_{\G'}
  \left(
  h(\mu_{k+1|k})^{-1}z_{k+1}
  \right)
  \right]^{\vee}_{\G'}\,.
  \end{align}
Finally, the update of the system state and covariance matrix can be evaluated as \cite{Bourmaud2014}
\begin{align}
  \mu_{k+1} & = \mu_{k+1|k} \exp_{\G}  \left( [\nu_{k+1}]^{\wedge}_{\G} \right) \label{eq:lgekf_mu_update}\\
  P_{k+1} & = \Phi_{\G}(\nu_{k+1}) 
  \left(
  \I^{p \times p} - K_{k+1} \H_{k+1}
  \right) P_{k+1|k}
  \Phi_{\G}(\nu_{k+1})^T \,.\label{eq:lgekf_cov_update}
\end{align}
We can notice similarities between the LG-EKF and EKF equations and, indeed, when $\G$ and $\G'$ are Euclidean spaces the LG-EKF reduces to EKF \cite{Bourmaud2014}.
Furthermore, due to the results \eqref{eq:adjoint_operators} and \eqref{eq:adjoint_operators_so2rr}, matrices
$\F_k$ and $\Phi_{\G}(\nu)$ for both $\SO(2)$ and $\SO(2)\times\R^2$ evaluate to
\begin{equation}\label{eq:lgekd_PhiG_Fk_SO2}
  \Phi_{\G}(\nu) = \I, \quad \F_k = \I + \mathscr{C}_k.
\end{equation}

In the sequel we derive the LG-EKF for the groups which we propose to utilize for tracking of moving objects with angular measurements and show that in this special case the LG-EKF reduces to the heuristically modified EKF.

\subsection{LG-EKF on $\SO(2)$}

In this section we derive the LG-EKF filter for state estimation on $\G=\SO(2)$.
For this group, mathematically dense LG-EKF equations are simplified and serve well to intuitively grasp the
mechanics of the filter.

\subsubsection{Prediction}
Let us take two examples of system models.
In the first we assume a stationary process, i.e., in the prediction the mean
value will remain unchanged except for the uncertainty that is added through the process noise (this is similar to the von Mises filter \cite{Azmani2009a})
\begin{equation}\label{eq:ekf_brown_system_model}
  x_{k+1|k} = x_k + n_k,  \quad n_k \sim \mathcal{N}_{\R^1} (0, \sigma_{Q}^2).
\end{equation}
This yields the LG-EKF system model $\Omega(X_k) = 0$ with the same process noise $n_k$,
which when inserted in \eqref{eq:lgekf_mu_prediction} will evaluate through the exponential as an identity matrix, thus
leaving the mean value unperturbed.

In order to compute the prediction of the covariance matrix via \eqref{eq:lgekf_cov_prediction}, given the result in \eqref{eq:lgekd_PhiG_Fk_SO2}, we only need to determine $\mathscr{C}_k$. 
In this case the Lie algebraic error is $\epsilon\in\R^1$ and due to the system model 
 the matrix $\mathscr{C}_k$ evaluates to zero, thus leaving $\F_k = 1$, and the prediction equations are
\begin{align}
  \mu_{k+1|k} = \mu_{k}, \quad P_{k+1|k} =  P_k + Q_k.
\end{align}
As we can see, these are the same formulae that an EKF prediction would yield with \eqref{eq:ekf_brown_system_model} as
the system model.

As the second example, we take the non-linear system \cite{Kurz2013a} where the robot rotary joint angle
was estimated
\begin{align}\label{eq:ekf_2_system_model}
  x_{k+1|k} = x_k + c_1 \sin(x_k) + c_2 + n_k,
\end{align}
where second and third term account for gravity and velocity, while the final term is again one-dimensional white
Gaussian noise.
This yields the following LG-EKF system model
\begin{align}\label{eq:so2_2_system_model}
  \Omega(X_k) &= c_1 \sin([\log(X_k)]^{\vee}_{\G}) + c_2.
\end{align}
Note that $[\log(X_k)]^{\vee}_{\G}$ is necessary to bring the rotation matrix with parameter $\mu_k$ to a scalar angle in $\R^1$.
The Lie algebraic error is again $\epsilon\in\R^1$ and given the system model \eqref{eq:so2_2_system_model} matrix
$\mathscr{C}_k$ evaluates to
\begin{align}\label{eq:lgekf_Ck_matrix_SO2}
  \mathscr{C}_k &= \dfrac{\partial}{\partial \epsilon} \Omega \left( 
  \begin{bmatrix}
  \cos\mu_k & -\sin\mu_k\\
  \sin\mu_k & \cos\mu_k
  \end{bmatrix}
  \begin{bmatrix}
  \cos\epsilon & -\sin\epsilon\nonumber\\
  \sin\epsilon & \cos\epsilon
  \end{bmatrix}
  \right)_{|\epsilon=0}\\
  &= \dfrac{\partial}{\partial \epsilon} \Omega \left(
  \begin{bmatrix}
    \cos(\mu_k + \epsilon) & -\sin(\mu_k + \epsilon)\\
    \sin(\mu_k + \epsilon) & \cos(\mu_k + \epsilon)
  \end{bmatrix}
  \right)_{|\epsilon=0}\nonumber\\
  &= \dfrac{\partial}{\partial \epsilon} \left( c_1 \sin(\mu_k + \epsilon) + c_2
  \right)_{|\epsilon=0}= c_1 \cos\mu_k
\end{align}
This means that $\F_k = 1 + c_1 \cos\mu_k $, and that the LG-EKF prediction equations are
\begin{align}
  \mu_{k+1|k} &= \mu_{k}\exp_{\G} \left( [c_1 \sin([\log(X_k)]^{\vee}_{\G}) + c_2]^{\wedge}_{\G}
  \right)\nonumber\\
   P_{k+1|k} &=  P_k (1 + c_1 \cos\mu_k )^2 + Q_k.
\end{align}
We can see that the covariance prediction formula is identical to the EKF covariance prediction.

More generally, to demonstrate the equivalence of the modified EKF and $\SO(2)$ LG-EKF prediction steps we need to
show that $\F_k = 1 + \mathscr{C}_k$ is equal to 
\begin{align}
F_k = \dfrac{\partial f_k(x_k, u_k)}{\partial x_k}_{|x_k = \mu_k},
\end{align}
where $F_k$ is the state transition matrix, i.e., the EKF system state Jacobian of \eqref{eq:kf_state_prediction}.
By inspecting \eqref{eq:lgekf_Ck_matrix_SO2} we can notice that for $\SO(2)$ the argument within $\Omega$ will always be
the sum of the mean value and the Lie algebraic error $\mu_k + \epsilon$.
This gives
\begin{align}
  \F_k &= 1 + \dfrac{\partial}{\partial \epsilon} \Omega(\exp_{\G}([\mu_k + \epsilon]^{\wedge}_{\G}),
  u_k)_{|\epsilon=0}\nonumber\\
  &= 1 + \dfrac{\partial}{\partial \epsilon} \hat{f}_k(\mu_k + \epsilon, u_k)_{|\epsilon=0}\nonumber\\
  &= 1 + \dfrac{\partial}{\partial \epsilon} (f_k(\mu_k + \epsilon, u_k) - (\mu_k +
  \epsilon_{\theta}))_{|\epsilon=0}\nonumber\\
  &= \dfrac{\partial f_k(\mu_k + \epsilon, u_k)}{\partial \epsilon}_{|\epsilon=0}
  = \dfrac{\partial f_k(\xi_k, u_k)}{\partial \xi_k}_{|\xi_k=\mu_k},
\end{align}
where variable substitution was performed in the last step: $\xi_k \leftarrow \mu_k + \epsilon, \partial
\xi_k \leftarrow \partial \epsilon$.
In the end $\F_k$ evaluates to the EKF Jacobian $F_k$ when the underlying group is $\SO(2)$.

\subsubsection{Correction}
Since we are measuring angles, we define the measurement Lie group as $\G'=\SO(2)$ and the measurement function
$h : \SO(2)\rightarrow\SO(2)$
\begin{equation}
  h(X_{k+1}) = R_{k+1}, \quad m_{k+1}\sim\mathcal{N}_{\R^1}(0, \sigma_R^2),
\end{equation}
which is trivial since the measurement and state group are the same, while the measurement noise is a
one-dimensional white Gaussian noise.
As in the prediction step, the associated Lie algebra is $\g'=\so(2)$.

To compute the correction step, we need to evaluate \eqref{eq:calH} for the LG-EKF on
$\SO(2)$.
The composition of the predicted mean $\mu_{k+1|k}$ and the Lie algebraic error yields
\begin{equation}\label{eq:lie_error_composition_prediction}
  \begin{split}
  h(\mu_{k+1|k}&\exp_{\SO(2)}([\epsilon]^{\wedge}_{\SO(2)}))
  =\\
  &\begin{bmatrix}
    \cos(\mu_{k+1|k} + \epsilon) & -\sin(\mu_{k+1|k} + \epsilon)\\
    \sin(\mu_{k+1|k} + \epsilon) & \cos(\mu_{k+1|k} + \epsilon)
  \end{bmatrix}.
  \end{split}
\end{equation}
Since $h(\mu_{k+1|k})^{-1}$ is the transpose of the corresponding rotation matrix, by inserting these results in
\eqref{eq:calH} we can calculate the measurement matrix
  \begin{align}
  \H_{k+1} &= \dfrac{\partial}{\partial \epsilon}\left( \left[ \log_{\G}
  \left(
  \begin{bmatrix}
    \cos\epsilon & -\sin\epsilon\\
    \sin\epsilon & \cos\epsilon
  \end{bmatrix}
  \right)
  \right]^{\vee}_{\G} \right)_{|\epsilon=0} = 1
  \end{align}
Given this results it is straightforward to see that Kalman gain and covariance update equation of $\SO(2)$ LG-EKF are
equal to the EKF equations.
The state correction equations yield the same result, except that the LG-EKF takes wrapping into account by composition
of rotation matrices, while the modified EKF computes everything in $\R^1$ and would need
to wrap the corrected state with $\wPi(\,\cdot\,)$.

\subsection{LG-EKF on $\SO(2)\times\R^2$}
\noindent In this section we derive the LG-EKF filter for state estimation on $\G=\SO(2)\times\R^2$.
Given the demonstrated equality of the $\SO(2)$ LG-EKF and the modified EKF and that LG-EKF reduces to EKF for Euclidean spaces, it is intuitive to expect that this result would extend to groups derived by composing $\SO(2)$ with Euclidean spaces.
In the sequel we illustrate this property by deriving a constant angular acceleration model for tracking with angle-only measurements.


\subsubsection{Prediction}

\noindent Given the state representation, we can now define the system model.
For this purpose, we use the constant angular acceleration model $\hat{\Omega}_k = \Omega(X_k):G\rightarrow\R^3$ 
\begin{equation}\label{eq:motion_model_so2rr}
  \Omega(X_k) =
  \begin{bmatrix}
    T\omega_{k} + \frac12 T^2 \alpha_k\\
    T\alpha_k\\
    0
  \end{bmatrix},
  \quad
  n_k \sim \mathcal{N}_{\R^3}(0, Q).
\end{equation}
Note that the displacement due to motion is calculated first in $\R^3$ and then according to
\eqref{eq:lgekf_mu_prediction} transferred to Lie algebra $\mathfrak{g}$,
exponentially mapped to the group $\G$ and then by way of composition added to the system state $X_k$.

In this case, the Lie algebraic error is $\epsilon = [\epsilon_{\theta} \ \epsilon_{\omega} \
\epsilon_{\alpha}]^{\text{T}}\in\R^3$, hence the composition of the mean value $\mu_k$ and $\epsilon$ yields
\begin{equation}
\begin{split}
  \mu_k\exp_{\G}([\epsilon]^{\wedge}_{\G})
  &= 
  \begin{pmatrix}
    R_kR_{\epsilon}\\
    \omega_k + \epsilon_{\omega}\\
    \alpha_k + \epsilon_{\alpha}
  \end{pmatrix}_{\G},\\
\end{split}
\end{equation}
with $R_kR_{\epsilon}$ has the same for as the matrix product in \eqref{eq:lie_error_composition_prediction}.
By applying the motion model \eqref{eq:motion_model_so2rr} on this results we get
\begin{equation}\label{eq:motion_model_so2rr_2}
  \Omega(\mu_k\exp_{\G}([\epsilon]^{\wedge}_{\G}))
  =
  \begin{bmatrix}
    T(\omega + \epsilon_{\omega}) + \frac12 T^2(\alpha + \epsilon_{\alpha})\\
    T(\alpha + \epsilon_{\alpha})\\
    0
  \end{bmatrix}.
\end{equation}
To compute the prediction step for the covariance matrix, we need to calculate matrix $\F_k$.
Since adjoint operators are trivial, using \eqref{eq:motion_model_so2rr_2} we calculate
\begin{align}
  F_k &= \I + \mathscr{C}_{k} = 
  \begin{bmatrix}
    1 & T &  \frac12 T^2\\
    0 & 1 & T\\
    0 & 0 & 1
  \end{bmatrix}.
\end{align}
%
%
We can see that the matrix $\F_k$ evaluates to the well known transition matrix of the classical EKF constant
acceleration motion model.

\subsubsection{Correction}
Since here we track moving objects by measuring angles, we define the measurement Lie group $\G' =
\SO(2)$ and the measurement function $h:\SO(2)\times\R^2 \rightarrow \SO(2)$
\begin{equation}\label{eq:so2rr_measurement_model}
  h(X_{k+1})
  =
  R_{k+1}, \quad
  m_{k+1} \sim \mathcal{N}_{\R^1}(0, \sigma_R^2)
\end{equation}
which in this case simply extracts the rotation matrix $R_{k+1}$ from $X_{k+1}$.
Calculation of the matrix $\H_{k+1}$ is the same as in \eqref{eq:calH}, except that $\epsilon$ is now a vector
\begin{equation}
  \begin{split}
  \H_{k+1} = 
  \dfrac{\partial}{\partial \epsilon} \left( \epsilon_{\theta} \right)_{| \epsilon=0} = [1 \ 0  \ 0].
  \end{split}
\end{equation}
Again, the same result as we would expect for the EKF measurement matrix.

\section{Experiments}

\begin{figure}[!t]
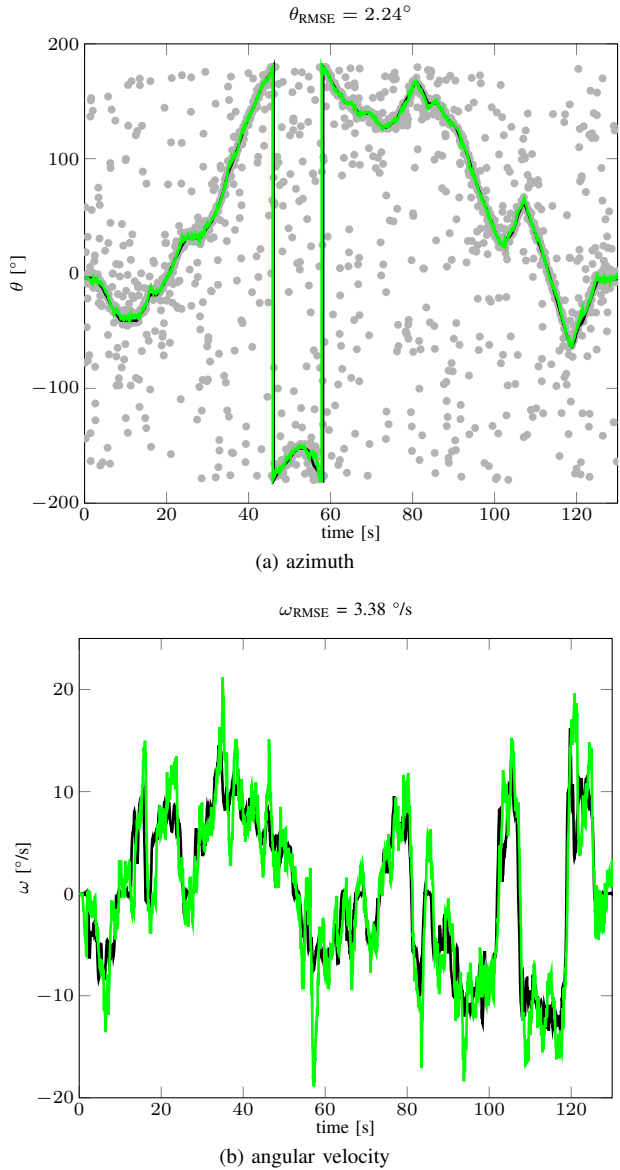

\centering
  \hspace{-0.5cm}
  \subfloat[azimuth]{
  \tikzsetnextfilename{so2rr_voice_angle}
  \input{figures/so2rr_voice_angle.tikz.tex}%
  \label{fig:so2rr_voice_angle}}\\
  \hspace{-0.5cm}
  \subfloat[angular velocity]{
  \tikzsetnextfilename{so2rr_voice_omega}
  \input{figures/so2rr_voice_omega.tikz.tex}%
  \label{fig:so2rr_voice_omega}}
  \caption{Performance of the LG-EKF on $\SO(2)\times\R^2$ when tracking a moving speaker. The solid black line is
  the ground truth as given by the motion capture system, the green solid line is the estimated state of the speaker,
  while the gray circles represent measurements, i.e. outputs of the beamformer. State RMSE is given in the title of each of the subfigures.
  }
  \label{fig:so2rr_voice}
\end{figure}


\noindent As a practical example of an application of the studied filter we apply this on the problem of speaker
tracking with a microphone array and in the present paper we test the $\SO(2)\times\R^2$ LG-EKF on real-world data.
For the sound acquisition we used the ManyEars framework consisting of an 8-channel USB sound card \cite{Grondin2013},
while for obtaining measurements we used the beamforming algorithm for speaker localization \cite{Valin2007} implemented
under the Robot Operating Systems \cite{Quigley2009} within the same framework.
The maximum of the beamforming energy was picked as the speaker measurement.

The experiments were conducted in a $120\,\text{m}^{2}$ room with parquet wooden flooring and one side covered with
windows.
The speaker was simulated by a loudspeaker playing an excerpt from Nature's podcast \emph{Audiophile} in English.
The area in which the loudspeaker moved was covered by a motion capture system, which was used to generate ground truth
data.
In order to handle outliers, we used validation gating; namely, the innovation matrix $S_{k+1} =
\H_{k+1}P_{k+1|k}\H^{\text{T}}_{k+1}+R_{k+1}$ was calculated and we applied the standard $\chi^2$--test
\begin{equation}
  \nu_{k+1}S^{-1}_{k+1}\nu_{k+1}^{\text{T}} < \gamma,
\end{equation}
where the threshold $\gamma$ was determined from the inverse $\chi^2_{p}$ cumulative distribution at a significance
level $a=0.95$ and $p$ degrees-of-freedom.
Figure~\ref{fig:so2rr_voice} shows the experiment results and corroborates that the filter successfully manages to
track the moving speaker in spite of the number of outliers.
Note that the modified EKF would yield the same results, except that in the case of the LG-EKF the system state was
defined on $\SO(2)\times \R^2$ and the idiosyncrasies of angular data were
intrinsically taken care of.

\balance

\section{Conclusion}
In this paper we have studied directional moving object tracking in 2D based on the extended Kalman filter
on matrix Lie groups.
First, we have proposed to analyze this estimation problem by modeling the state to reside on the $\SO(2)$ group.
Subsequently, we have shown that the $\SO(2)$ filter derivation based on the mathematically grounded framework of
filtering on Lie groups yields evaluates to heuristically wrapping the extended Kalman filter.
We emphasize that this result \emph{applies only} to the $\SO(2)$ filter and is not intended to be extended to other
Lie groups or combinations thereof.
Second, we have derived the constant angular velocity $\SO(2)\times\R^2$ filter, where the system state consisted
of azimuth, angular velocity and angular acceleration.
For this filter we showcased a real-world experiment of a speaker tracking problem with a microphone
array by assessing the accuracy using the ground truth obtained by a motion capture system.

\section*{Acknowledgments}
This work has been supported from the European Union’s Horizon 2020 research and innovation programme under grant
agreement No 688117 (SafeLog) and the Unity Through Knowledge Fund under the project Cooperative Cloud based
Simultaneous Localization and Mapping in Dynamic Environments.

\bibliographystyle{IEEEtran}
\bibliography{bibliography/EstimationManifold,bibliography/bibliography,bibliography/library}

\end{document}